\documentclass[fleqn,10pt]{wlscirep}
\usepackage[utf8]{inputenc}
\usepackage[T1]{fontenc}
\usepackage{amsmath}
\usepackage{amsthm}
\usepackage{amssymb}
\usepackage{lipsum}
\usepackage{subfigure}
\usepackage{graphicx}
\usepackage{makecell}
\usepackage{verbatim}
\usepackage{hyperref}
\usepackage{threeparttable}
\usepackage{breakurl}
\usepackage{multirow}

\title{{\color{black} A Satellite Imagery Dataset for Long-Term Sustainable Development in United States Cities}}

\author[1]{Yanxin Xi}
\author[2,3]{Yu Liu}
\author[2,3]{Tong Li}
\author[2,3]{Jintao Ding}
\author[2,3]{Yunke Zhang}
\author[1]{Sasu Tarkoma}
\author[2,3,*]{Yong Li}
\author[1,4,*]{Pan Hui}
\affil[1]{Department of Computer Science, University of Helsinki, Helsinki, Finland}
\affil[2]{Beijing National Research Center for Information Science and Technology (BNRist), Beijing, P. R. China}
\affil[3]{Department of Electronic Engineering, Tsinghua University, Beijing, P. R. China}
\affil[4]{Department of Computer Science and Engineering, Hong Kong University of Science and Technology, Hong Kong, P. R. China}
\affil[*]{corresponding author(s): Yong Li (liyong07@tsinghua.edu.cn) and Pan Hui (pan.hui@helsinki.fi)}

\begin{abstract} 
Cities play an important role in achieving sustainable development goals (SDGs) to promote economic growth and meet social needs. Especially satellite imagery is a potential data source for studying sustainable urban development. {\color{black}However, a comprehensive dataset in the United States (U.S.) covering multiple cities, multiple years, multiple scales, and multiple indicators for SDG monitoring is lacking. To support the research on SDGs in U.S. cities, we develop a satellite imagery dataset using deep learning models for five SDGs containing 25 sustainable development indicators. The proposed dataset covers the 100 most populated U.S. cities and corresponding Census Block Groups from 2014 to 2023}. Specifically, we collect satellite imagery and identify objects with state-of-the-art object detection and semantic segmentation models to observe cities’ bird’s-eye view. {\color{black}We further gather population, nighttime light, survey, and built environment data to depict SDGs regarding poverty, health, education, inequality, and living environment.} We anticipate the dataset to help urban policymakers and researchers to advance SDGs-related studies, especially applying satellite imagery to monitor long-term and multi-scale SDGs in cities.
\end{abstract}
\begin{document}

\flushbottom
\maketitle

\thispagestyle{empty}


\section*{Background \& Summary}
Nowadays, more than $50\%$ of the population lives in cities, producing $80\%$ of the GDP worldwide\cite{Urban2022, World2022}. Therefore, cities play an increasingly important role in achieving United Nations Sustainable Development Goals\cite{Transforming2015} (SDGs), which aim to prosper economic growth and meet social needs. {\color{black} According to the report "The United States Sustainable Development Report\cite{UNSDSN}" from the United Nations, cities in the United States (U.S.) perform poorly on a series of SDGs (e.g., Boise city lags behind in quality education, and Raleigh city shows high poverty rate\cite{2017SDSN}).}
{\color{black} Currently, monitoring sustainable development in U.S. cities heavily relies on door-to-door surveys such as the American Community Survey\cite{ACS,SG} (ACS) data. First, ACS data for constructing the SDG index in U.S. cities is economically costly as the annual budget can reach millions of dollars\cite{ACSBudget}. Second, the current SDG index dataset for U.S. cities is meant for a single year and only focuses on the city level, which hinders monitoring of multi-scale and multi-year SDG progress\cite{SDSN}.}
Alternatively, built upon the rapid development of remote sensing and deep learning techniques, satellite imagery showing nearly real-time and bird's-eye view information in cities has been broadly investigated as a data source for SDG monitoring \cite{burke2021using,jean2016combining,head2017can,chen2020analysis,ayush2020generating,albert2017using,wang2018urban,yeh2020using}. Therefore, monitoring SDGs in cities with satellite imagery is of great significance in promoting sustainable urban development.

{\color{black}However, a long-term and multi-scale satellite imagery dataset, which reveals the yearly change in SDGs in multiple years and in different spatial (administrative) scales for city SDG monitoring, is still lacking. For instance, some satellite imagery datasets about SDGs focus on either country level or cluster level (25-30 households) spatially and only contain data of a single year \cite{sumbul2019bigearthnet}, which barely match the requirements of long-term and multi-scale SDG monitoring in cities.} Other open-source satellite imagery datasets, such as SpaceNet\cite{van2018spacenet} or ForestNet \cite{irvin2020forestnet}, merely contain the dataset for one single SDG. {\color{black} Besides, other survey data for SDG, such as UNESCO Survey on Public Access to Information and Survey Module \cite{UNESCO} on SDG Indicator 16.b.1 \& 10.3.1, are based on questionnaires.  At last, although plenty of survey data in the U.S. will aid in SDG monitoring, it would be difficult for urban policymakers and researchers to extract the critical information easily. Motivated by the SustainBench\cite{yeh2021sustainbench} in low-income countries, and to fulfill the data requirements for SDG monitoring in U.S. cities, we propose a comprehensive long-term and multi-scale satellite imagery dataset with {\color{black} 25} SDG indicators for {\color{black}five} SDGs (SDG 1, SDG 3, SDG 4, {\color{black}SDG 10}, and SDG 11). Moreover, the dataset covers about 45,000 Census Block Groups (CBGs) in the 100 most populated U.S. cities from 2014 to 2023}. {\color{black} Using satellite images and SDG data in the U.S., urban policymakers and researchers can develop various models or assumptions regarding SDG monitoring remotely. And further, the dataset from the U.S. can aid urban policymakers and researchers in inferring SDG progress in low-income countries, which mostly lack SDG-related survey data.}

\begin{figure}[bh]
\centering
\includegraphics[width=5in,trim={0 0 0 0},clip]{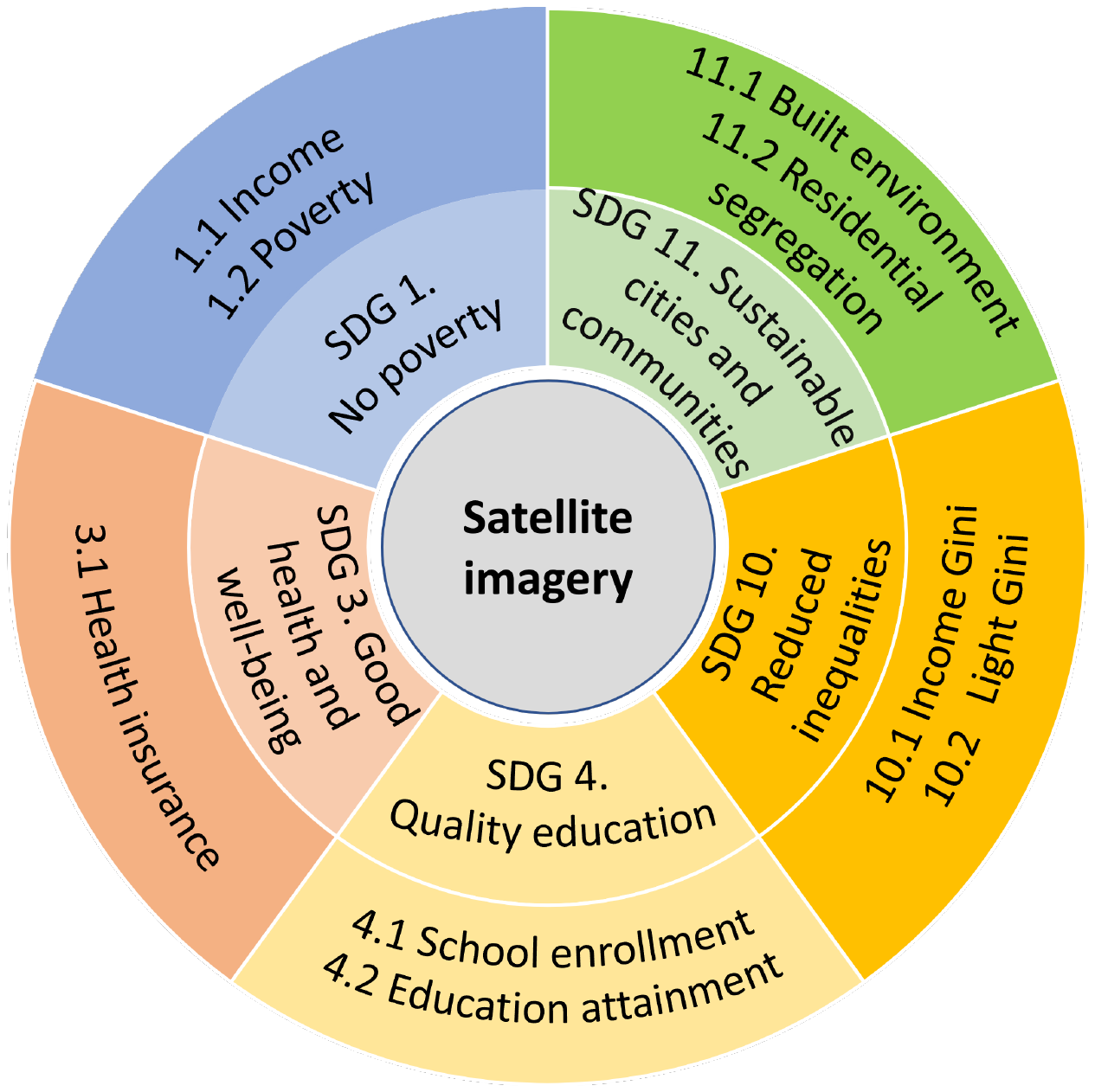}
\caption{{\color{black}Schemetic overview of the target dataset.}}
\label{fig:overview}
\end{figure}

{\color{black} Figure \ref{fig:overview} presents the scheme of our produced dataset with two components: the satellite imagery data containing the detected objects and land cover semantics obtained with state-of-the-art deep learning models and the corresponding SDG indicators in 100 cities from 2014 to 2023}. For the satellite imagery data, we consider the daytime satellite imagery with the spatial resolution of $0.3$m and several objects (such as truck and basketball court\cite{xview,ding2019learning,xia2018dota,Ding2021}) detected as well as several land cover semantics (forest and road\cite{wang2021loveda}) inferred with the models transferred from the computer vision community. The detected objects refer to the countable artificial objects and venues visible in satellite images in cities, while the land cover is mostly the uncountable environmental information. For SDGs, we collect the indicators that can be inferred from satellite images in urban scenarios. {\color{black} Specifically, indicators for SDG 1 "No poverty", SDG 3 "Good health and well-being", SDG 4 "Quality education", SDG 10 "Reduced inequalities", and SDG 11 "Sustainable cities and communities" \cite{Transforming2015,guenat2022meeting} are included in our dataset.} The indicators are generated from multi-source data, including nighttime light (NTL) data from Earth Observation Group (EOG) \cite{elvidge2017viirs}, WorldPop population data\cite{tatem2017worldpop}, ASC data\cite{ACS,SG}, and OpenStreetMap (OSM) built infrastructure \cite{OSM20221,OSM20222}. 


{\color{black} At last, this paper advances the SDG-related community by generating a long-term and multi-scale satellite imagery dataset in urban scenarios by collecting and processing satellite images and SDG indicators from multiple sources, which is a time-consuming and laborious work, and the alignment of satellite image visual attributes and SDG data. The dataset aims to help urban policymakers and researchers, who might not have the platform to collect and process the large volume of data, to conduct numerous SDG studies spanning poverty, health, education, inequality, and built environment. More importantly, as the first urban satellite imagery dataset for multiple SDGs monitoring with interpretable visual attributes (e.g., cars, buildings, roads, etc.), it can aid in further achieving sustainable cities with high interpretability and advancing urban sustainability progress.} The satellite imagery and the visual attributes extracted by the computer vision models in the dataset can serve as the input for various kinds of research regarding SDGs, and the SDG indicators act as output. Specifically, we recommend the following potential applications:
\begin{itemize}
    \item Researchers can design deep learning models to predict various long-term SDGs (income, poverty, and built environment) in cities from historical satellite images.
    \item Researchers can also estimate various SDG progresses by utilizing multi-scale satellite imagery visual attributes at the CBG and city levels and reveal the linkage between the multi-scale satellite imagery and SDGs.
    \item Researchers can propose a spatiotemporal framework that simultaneously utilizes long-term and multi-scale satellite imagery for SDG monitoring, which sheds light on satellite imagery fusion of temporal and spatial dimensions.  
\end{itemize}




\section*{Methods}
\begin{figure}[tb]
\centering
\includegraphics[width=7in]{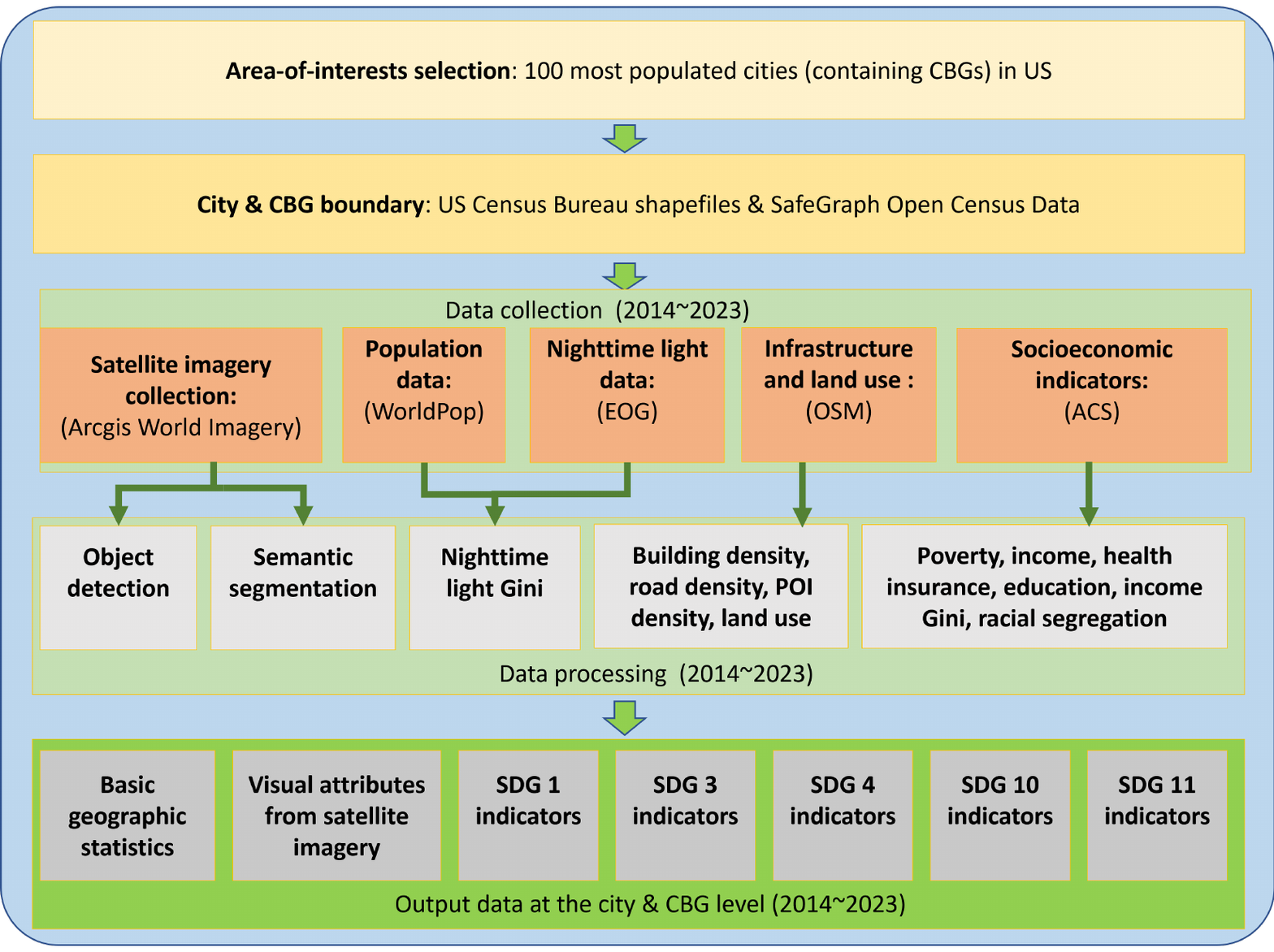}
\caption{{\color{black}Overall workflow of the dataset generation process.}}
\label{fig:workflow}
\end{figure}

We aim to provide a comprehensive and representative dataset that includes satellite imagery and corresponding SDG indicators covering long terms and multiple scales. {\color{black} To ensure that the indicators can thoroughly depict sustainable urban development, we select five SDGs altogether: SDG 1 No poverty (five indicators), SDG 3 Good health and well-being (five indicators), SDG 4 Quality education (five indicators), SDG 10 Reduced inequalities (two indicators), and SDG 11 Sustainable cities and communities (eight indicators).}
Overall, the target dataset generation process includes collecting, processing, and aligning multi-source data, and the overall workflow is presented in Figure \ref{fig:workflow}. We first select the 100 most populated cities and gather the corresponding CBG/city boundaries. Second, we collect satellite imagery, population, NTL, OSM, and ACS data from multiple sources. At last, we process the multi-source data and produce the final output data at the CBG and city levels, containing basic geographic statistics, satellite imagery attributes, and SDG indicators. 


\subsection*{Determining the area-of-interests and boundary extraction} \label{bound}
We select the 100 cities with the most population in the contiguous United States, which is explored on the ACS 2021 population data\cite{citylist}. The population in the 100 cities varies from 222,194 to 8,467,513, with a mean population of 642,002. The city-of-interests and population in descending order are shown in Table \ref{cityname}.

Then we collect city geographic boundary files from the U.S. Census Bureau TIGER/Line shapefiles \cite{citybound}. {\color{black} The shapefiles are divided by states, and each shapefile contains the city name (called "place" in the file), state name, Federal Information Processing Standard state code, and geographical boundary coordinates.} We use the python packet \textit{shapely} to access the shapefiles and extract the boundary coordinates using the city and state names. The geographic coordinate system is WGS84.

{\color{black} Next, we determine the corresponding CBGs within the cities. The boundaries of all CBGs in the U.S. are gathered from SafeGraph Open Census Data \cite{SG}. The CBG boundaries for the years 2014 \textasciitilde \ 2019 are the same, and the U.S. government adjusts the CBG boundary for the year 2020.
For each city, we overlap the CBG boundaries on the city boundary, and every CBG whose area intersection with city boundary takes up more than $10\%$ of the corresponding CBG area is considered contained in the city. This process uses Python packets \textit{shapely} and \textit{geopandas}. The geographic lookup table between cities and corresponding CBGs is shown in Table \ref{tab2}. Till this step, we have the selected 100 most populated cities and corresponding CBGs spatially contained as the area-of-interests in our target dataset.}

\subsection*{Processing of satellite imagery}
{\color{black} Satellite imagery provides a near real-time bird's-eye view of the earth's surface. Combined with machine learning techniques, satellite imagery has been widely used in predicting socioeconomic status, especially in urban research, which includes poverty/asset prediction\cite{ayush2020generating,yeh2020using,jean2016combining}, urban pattern mining\cite{albert2017using}, commercial activity prediction\cite{wang2018urban,10.1145/3184558.3186353}, and population prediction\cite{han2020learning,head2017can}. Inspired by the interpretable feature generation from satellite imagery\cite{ayush2020generating}, we provide satellite imagery visual attributes in our dataset to promote the research of SDG monitoring.} The processing of satellite imagery consists of three parts: imagery collection, object detection, and semantic segmentation.

First, we collect the satellite imagery in our dataset from Esri World Imagery \cite{esri}. {\color{black} It provides users access to the World Imagery of different versions created over time. The imagery is in RGB format collected from different satellites and of different spatial resolutions marked by different zoom levels, which split the entire world into different numbers of tiles.}
Overall, the imagery collection process includes generating image tile numbers according to the boundary of each city as well as the desired zoom level (spatial resolution) and downloading images with the tile numbers from the satellite imagery archive. In our target dataset, we set the zoom level to $19$, which is about $0.3$m/pixel. {\color{black} We also select the Esri World Imagery archive of June from 2014 to 2023 to collect the satellite images of the 100 most populated cities, which generates altogether 12,269,976 images each year.}
{\color{black} Second, many aspects of cities are related to people's lives and can reveal SDG progress. Transportation in the city is integral to urban development \cite{knowles2020transport}, and further, transportation and mobility were recognized as central to sustainable development at the 2012 United Nations Conference on Sustainable Development\cite{sustainableTransport}. Sports \& leisure are highly correlated to citizens' life quality\cite{carlucci2020socio, wang2019influence}. Children and young people benefit largely from sports, which are inseparable from a quality school education, promoting SDG 3 and SDG 4\cite{sport}. The building  {\color{black} characteristics} (e.g., building type) can reveal the population and income status in urban areas\cite{abitbol2020interpretable,chen2021uvlens}, and the impact of buildings on human well-being can not be neglected. Therefore, the buildings, cars, and other objects in satellite imagery contain certain correlations with SDG indicators. In our dataset, we consider 17 objects from the abovementioned aspects: transportation, sports \& leisure, and building.}

The urban object categories are presented in Table \ref{tab:obdss}. {\color{black} We use the YOLOv5s model\cite{glenn_jocher_2022_7347926,yolov5code} pre-trained on the MS COCO dataset \cite{lin2014microsoft} and finetune it on xView dataset \cite{xview} and DOTA v2 dataset \cite{xia2018dota} to detect objects in the collected satellite imagery. The default parameters\cite{yolov5param} are used for finetuning the object detection models.} We aggregate the number of objects detected from satellite images at the CBG and city levels to show visual object attributes at multiple scales.

Third, land cover information such as forests or water can also depict the urban environment and is not included in the detected objects. Therefore, we add the land cover semantic information inferred from satellite imagery in our generated dataset. {\color{black} We use the Vision Transformer (ViT)-Adapter-based semantic segmentation model \cite{chen2022vitadapter, ViT,vitparam} pre-trained on the ADE20K dataset\cite{zhou2017scene} and finetune it on LoveDA dataset \cite{wang2021loveda} to generate semantic information from the collected satellite imagery, which includes background, building, road, water, barren, forest, and agriculture. Moreover, we compute the pixel-level percentage of each semantic information presented in Table \ref{tab:obdss} in each satellite imagery and aggregate them at the CBG and city levels, respectively.}

\subsection*{Processing of basic geographic statistics}
For each CBG/city, we present the population, area, centroid coordinates, and geographic boundary, which describe the essential information for the selected area-of-interests. Specifically, we collect the population data from 2014 to  {\color{black}2020} from the WorldPop project \cite{tatem2017worldpop,WP}. The population data is downloaded at a resolution of 3 arc (approximately $100$m at the Equator). We use Python packets \textit{shapely} and \textit{gdal} to crop the population data with the CBG/city geographic boundary and sum up the cropped pixel values as the total population. The area ($km^2$) is calculated from the CBG/city boundary data with Python packet \textit{geopandas}. The geographic centroid can also be computed with Python packet \textit{geopandas}.


\subsection*{Processing of SDG indicators}
{\color{black} There are five SDGs (SDG 1, SDG 3, SDG 4, SDG 10, and SDG 11) concerning poverty, health, education, inequality, and built environment collected in our produced dataset at the CBG/city level.} SDG 1 "No poverty" focuses on income and population in poverty status. The indicators for "No poverty" are collected from ACS data. SDG 3 "Good health and well-being" and SDG 4 "Quality education" highlight people's health insurance status and population with different academic degrees, and corresponding indicators are extracted from ACS data. {\color{black} SDG 10 (Reduced inequalities) intends to reduce inequality, and the indicators are from ACS data and from NTL combined with population data with a recent algorithm for monitoring regional inequality through NTL\cite{mirza2021global}.} Finally, SDG 11 "Sustainable cities and communities" reflects the living conditions in CBG/city, and the related indicators are calculated from OSM historical data and ACS data. {\color{black}Altogether, we collect $25$ indicators across five SDGs. 
The indicators and relevant SDG targets are described in Table \ref{tab_corr}.}

\paragraph{Indicators for SDG 1 "No poverty".}
SDG 1 aims to end poverty in all its forms everywhere \cite{Transforming2015}. Our target dataset incorporates income and poverty status data to represent the SDG 1 indicators in cities. Specifically, median household income, population above poverty (number of population whose income in the past 12 months is at or above poverty level), population below poverty (number of population whose income in the past 12 months is below poverty level), and population with a ratio of income to poverty level (the total income divided by poverty level) under 0.5 and between 0.5 to 0.99 are collected to describe the income \& poverty in CBG/city. {\color{black} The poverty threshold is computed by the Census Bureau according to the family size and ages of family members every year with variations to Consumer Price Index. The threshold is a country-specific value and does not change geographically\cite{PovertyThresh}.} Moreover, population above/below poverty and population with different ratios of income to poverty level are measurements of poverty status.

We collect the median household income, population above/below poverty, and population with a ratio of income to poverty level under 0.5 and between 0.5 to 0.99 at the CBG level from the ACS data\cite{ACS,acs-guide,SG}. 
Then, we generate the city-level indicators: population above/below poverty and population with a ratio of income to poverty level under 0.5 and between 0.5 to 0.99 by aggregating all the CBG data within the city. Median household income at the city level is related to the income distribution of the population in cities and is gathered directly from ACS data\cite{B19013}.
{\color{black} The boundary files and ACS data are both collected from the U.S. Census Bureau. And ACS data denotes the city as "place" as in the boundary files, and the ACS definition of a city boundary is the same as the U.S. Census Bureau TIGER/Line shapefiles.}

\paragraph{Indicators for SDG 3 "Good health and well-being".}
SDG 3 aims to ensure healthy lives and promote well-being for all populations at all ages\cite{Transforming2015}. {\color{black}In our target dataset, we use the population data with no health insurance covering all ages to represent SDG 3 indicators because health insurance is correlated to the health status of the population in urban regions\cite{pan2016health,meng2018impact}.} Specifically, civilian noninstitutionalized population, population with no health insurance under 18, between 18 to 34, between 35 to 64, and over 65 years old are collected from ACS data\cite{SG} to describe the health insurance at the CBG and city levels.

\paragraph{Indicators for SDG 4 "Quality education".}
SDG 4 aims to ensure inclusive and equitable quality education and promote lifelong learning opportunities for all\cite{Transforming2015}. Therefore, indicators directly depicting city education status can be selected here. In dataset generation, we collect from ACS data\cite{SG} population enrolled in college, population that graduated from high school, population with a bachelor's degree, a master's degree, and a doctorate for indicators of school enrollment \& education attainment to monitor SDG 4. 

{\color{black}\paragraph{Indicators for SDG 10 "Reduced Inequalities".}}
{\color{black} SDG 10 aims to reduce inequality within and among countries\cite{Transforming2015}. We use income Gini\cite{damgaard2000describing} and light Gini\cite{mirza2021global} to monitor the process of SDG 10. The income Gini reveals the inequality status of income and is collected from ACS data. Light Gini can present the distribution of NTL per person and thus indirectly reveal regional development inequality. Similar to the income Gini, the lower light Gini is, the more equally the region develops, which means the region moves towards eliminating inequality in SDG 10. The results in the original paper\cite{mirza2021global} report the light Gini at a 1-degree grid cell, which can not be directly used in urban scenarios. Therefore, we calculate the light Gini following the method\cite{mirza2021global}. Specifically, the NTL per person is calculated by dividing the NTL value by the population number in all grids in each CBG/city. Then, the Gini index\cite{damgaard2000describing} of NTL per person in the CBG/city boundary is computed as the light Gini. The NTL is the Visible Infrared Imaging Radiometer Suite (VIIRS) data\cite{elvidge2017viirs,mirza2021global} with a spatial resolution of 15 arc seconds (500 m at the Equator). We download the VIIRS Nighttime Lights version 2 Median monthly radiance (the unit for light intensity is nW /cm2/sr) with background masked from EOG\cite{elvidge2013viirs,elvidge2017viirs,elvidge2021annual,EOG}. Compared with income Gini from traditional income survey data, light Gini measures the NTL inequality in urban regions by considering NTL as an indicator for economic development, which is a different measurement of inequality\cite{mirza2021global}.}

\paragraph{Indicators for SDG 11 "Sustainable cities and communities".}
SDG 11 aims to make cities and human settlements inclusive, safe, resilient, and sustainable \cite{Transforming2015}. {\color{black} We incorporate indicators related to the built environment and land use in the target dataset. Specifically, we generate building density, driving/cycling/walking road density, POI density, land use information, and residential segregation (index of dissimilarity and entropy index) as indicators to monitor SDG 11.}

The source data of urban built environment and land use is collected from OSM \cite{OSM20221,haklay2008openstreetmap,vargas2020openstreetmap}. We collect the U.S. state-level historical Protocolbuffer Binary Format files from Geofabrik\cite{OSM20222} from 2014 to 2023.
Then we apply Python packet \textit{pyrosm} to extract the building, driving road, cycling road, walking road, POI, and land use information in cities and CBGs by corresponding boundary polygons. For calculating building density, we divide the number of buildings by the area of CBG/city. For each of the three kinds of road density, we divide the total length of each kind of road by the corresponding area of CBG/city. The POI density, which is defined as the ratio of the number of all POIs and the area of CBG/city, can show urban venues with human information. {\color{black} The OSM POIs include all OSM elements with tags "amenity", "shop" or "tourism". The amenity tag is useful and important facilities for the urban population, which include Sustenance, Education, Transportation, Financial, Healthcare, Entertainment, Arts \& Culture, Public Services, Waste Management, and Others. The shop tag includes locations of all kinds of shops and the sold products, such as Food \& Beverages, General Store, Mall, Clothing, Shoes, Accessories, Furniture, etc. The tourism tag is the places for tourists, such as Museum, Gallery, Theme Park, Zoo, etc.}
Moreover, we generate the land use indicators (commercial, industrial, construction, and residential) by calculating the area percentage of each kind of land use in the area of CBG/city. 

{\color{black} The indicators for the built environment quantitatively measure the density of buildings and roads. It should be noted that the indicators for SDG 11 are imperfect since the actual quality of buildings and roads is not provided in the dataset. Users can use the building/road/POI indicators as side information for depicting urban development.}

{\color{black} Residential segregation is related to inclusivity in U.S. cities\cite{chodrow2017structure}. We calculate the index of dissimilarity \cite{sakoda1981generalized}
\begin{equation}
    D = \frac{1}{2}\sum^n_{i=1}\left|\frac{w_i}{w_T}-\frac{b_i}{b_T}\right|,
\end{equation}
where $n$ is the number of CBGs in a city, $w_i$ is the number of race "w" (e.g., White) in CBG $i$, $w_T$ is the total number of race "w" in the city, $b_i$ is the number of race "b" (e.g., Black) in CBG $i$, and $B_T$ is the total number of race "b" in the city. We calculate the index of dissimilarity for four racial or ethnic groups: Non-Hispanic White (White), Non-Hispanic Black or African American (Black), Non-Hispanic Asian (Asian), and Hispanic\cite{chodrow2017structure}. There are altogether six categories of indices of dissimilarity: White-Black, White-Asian, White-Hispanic, Black-Asian, Black-Hispanic, and Asian-Hispanic. 

Next, we calculate the entropy index \cite{iceland2004multigroup}
\begin{equation}
    h_i = -\sum^k_{j=1}p_{ij}ln(p_{ij}),
\end{equation}
where $k$ is the number of racial/ethnic groups, $p_{ij}$ is the proportion of $j^{th}$ race/ethnicity in CBG/city $i$. We include groups of the White, Black, Asian, and Hispanic population at the CBG or city level.}

{\color{black}\subsection*{Limitations}
The limitations of our dataset include errors from multiple data sources, partial coverage of SDG progress, and the shortcomings of selected indicators.  

The errors from data sources include measurement errors in satellite imagery, ACS data collection, OSM, WorldPop population, and NTL data. The measurement errors in satellite imagery processing are mainly from the object detection and semantic segmentation tasks, which are shown in Table \ref{tab_IOU}. While the errors in other data sources are usually tolerable in each field and the quality assessment can be referred to literature\cite{MOE} for ACS data, literature\cite{barrington2017world,zhou2018exploring,zhang2019using,zhou2022exploring} for OSM, literature\cite{ma2021accuracy} for WorldPop, and literature\cite{elvidge2021annual} for NTL data. ACS data uses sampling error to measure the difference between the true values for the entire population and the estimate based on the sample population. And the magnitude of sampling error is measured by the margin of error\cite{MOE}. ACS provides a margin of error for all ACS estimate data which we collect as SDG indicators in our dataset. The dataset users can freely access the margin of error values of the ACS-oriented indicators in our dataset from the ACS official website. OSM data is a Volunteered Geographic Information (VGI) and is frequently updated by volunteers. In terms of the road network, OSM is about $83\%$ complete globally\cite{barrington2017world}. The building completeness for OSM in San Jose city in the U.S. is about $72\%$ and confirms the validity of OSM building density in our dataset\cite{zhou2018exploring}. Some cities show a large jump in building number in a consecutive year due to lagging annotations. The POIs in OSM are compared with the Foursquare POIs and $60\%$ of the POIs can be matched with high accuracy\cite{zhang2019using}. At last, the accuracy of the OSM land use dataset\cite{zhou2022exploring} for the U.S. is above $60\%$. The population data from WorldPop has a coefficients of determination\cite{lewis2015applied} $R^2$ greater than 0.95 when evaluated on the population data in China\cite{ma2021accuracy}. The nighttime light intensity also shows a high consistency ($R^2$ greater than 0.97) compared with different nighttime light datasets\cite{elvidge2021annual}.

And the provided dataset does not cover the whole SDG aspects, and thus cannot be used as the sole measurement for SDG monitoring. However, the dataset still has great reference value and aids decision-making for urban researchers and policymakers. 

At last, some indicators cannot always be the best indicators for corresponding SDGs. For example, the indicator health insurance for SDG 3 (Good health and well-being) may not be the best measurement of health status because health insurance usage is affected by the income or wealth of the insurance owners.}







 

\section*{Data Records}
{\color{black} The produced dataset can be accessed through the Figshare repository\cite{OurData} and is stored in tabular format. We split the output dataset into seven categories, as shown in Figure \ref{fig:workflow}: basic geographic statistics, satellite imagery attributes, and five SDGs described in Figure \ref{fig:overview} to help users quickly access and utilize the data.} Moreover, for each category, the dataset also contains records at two spatial levels (city and CBG). First, to help the users understand the area-of-interests, we provide samples of the geographic lookup table between the cities and CBGs in Table \ref{tab2}, and the basic statistics of CBGs and cities in Table \ref{tab3}-\ref{tab4}. Second, to demonstrate the extracted visual attributes from the satellite imagery, we show samples of objects detected and land cover semantics from the satellite imagery at the CBG level in Table \ref{sat}. {\color{black} At last, the samples of SDG indicators at the CBG level are demonstrated in Table \ref{tab5}-\ref{tab9}, respectively, which include SDG 1 "No poverty", SDG 3 "Good health and well-being", SDG 4 "Quality education", SDG 10 "Reduced inequalities", and SDG 11 "Sustainable cities and communities".} The city name and CBG code are used to mark the geographical location of each SDG indicator. 

\subsection*{Data Table Formats}
\paragraph{Basic geographical statistics.}
Table \ref{tab3}-\ref{tab4} provide the population, area, geographic centroid, and geographic coordinate boundary of the area-of-interests in this dataset, where the area, geographic centroid, and boundary are invariant to time, while the population is time-varying.
\paragraph{Satellite imagery attributes.}
{\color{black}We have object numbers and land cover semantic attributes processed from satellite imagery of the years 2014 to 2023}. The object categories include planes, airports, passenger vehicles, trucks, railway vehicles, ships, engineering vehicles, bridges, roundabouts, vehicle lots, swimming pools, soccer fields, basketball courts, ground track fields, baseball diamonds, tennis courts, and buildings (number of buildings). The land cover semantic attributes contain background, building (pixel percentage), road, water, barren, forest, and agriculture. There are altogether 24 visual attributes obtained from satellite imagery, which are shown in Table \ref{sat}. For visualization convenience, we only show the samples at the CBG level.

\paragraph{SDG 1.}
{\color{black}We provide median household income, population above/below poverty, population with a ratio of income to poverty level under 0.5, and  population with a ratio of income to poverty level between 0.5 to 0.99 for "No poverty" indicators in Table \ref{tab5} for the years 2014 to 2023 at the CBG level.} 
\paragraph{SDG 3.}
{\color{black}We offer civilian noninstitutionalized population, population with no health insurance under 18, population with no health insurance between 18 to 34, population with no health insurance between 35 to 64, and population with no health insurance over 65 years old for "Good health and well-being" indicators in Table \ref{tab6} for the years 2014 to 2023 at the CBG level.} 
\paragraph{SDG 4.}
{\color{black} We provide population enrolled in college, population that graduated from high school, population with a bachelor's degree, a master's degree, and a doctorate for "Quality education" indicators in Table \ref{tab7} for the years 2014 to 2023 at the CBG level.} 

{\color{black}\paragraph{SDG 10.}
We provide income Gini and light Gini for "Reduced inequalities" indicators in Table \ref{tab8} for the years 2014 to {\color{black}2023} at the CBG level. The income Gini measures the regional inequality from the perspective of income, and the light Gini shows the regional nighttime light inequality through remote sensing technology.}


\paragraph{SDG 11.}
{\color{black}We provide building density, driving/cycling/walking road density, POI density, land use information (commercial, industrial, construction, and residential), and {\color{black}residential segregation (index of dissimilarity and entropy index)} for "Sustainable cities and communities" indicators in Table \ref{tab9} for the years 2014 to 2023 at the CBG level.}

\section*{Technical Validation}

\begin{figure}[tb]
\centering
\includegraphics[width=4in]{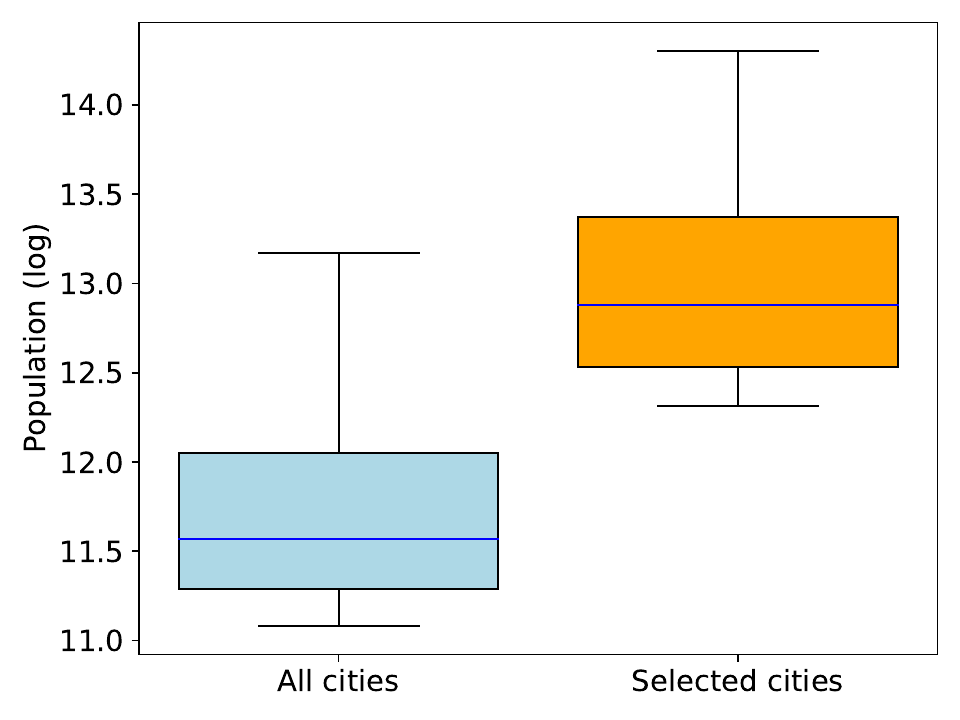}
\caption{{\color{black}Comparison of the overall population in city-of-interests in our dataset and population in all U.S. cities.}}
\label{FigRep}
\end{figure}

{\color{black}\subsection*{Population percentage of city-of-interests}}
Our dataset selects the 100 most populated cities in the contiguous United States. {\color{black}We demonstrate the comparison of overall population in our city-of-interests and in all U.S. cities\cite{citylist} in Figure \ref{FigRep}.} We find that the population in selected cities takes up $52\%$ of the total population in U.S. cities.

\subsection*{Visual attributes extraction from satellite imagery}
\begin{figure}[tb]
\centering
\subfigure[{\color{black}Object detection}]{
    \begin{minipage}[t]{0.45\linewidth}
        \centering
        \includegraphics[height=3in,trim={60 20 20 0},clip]{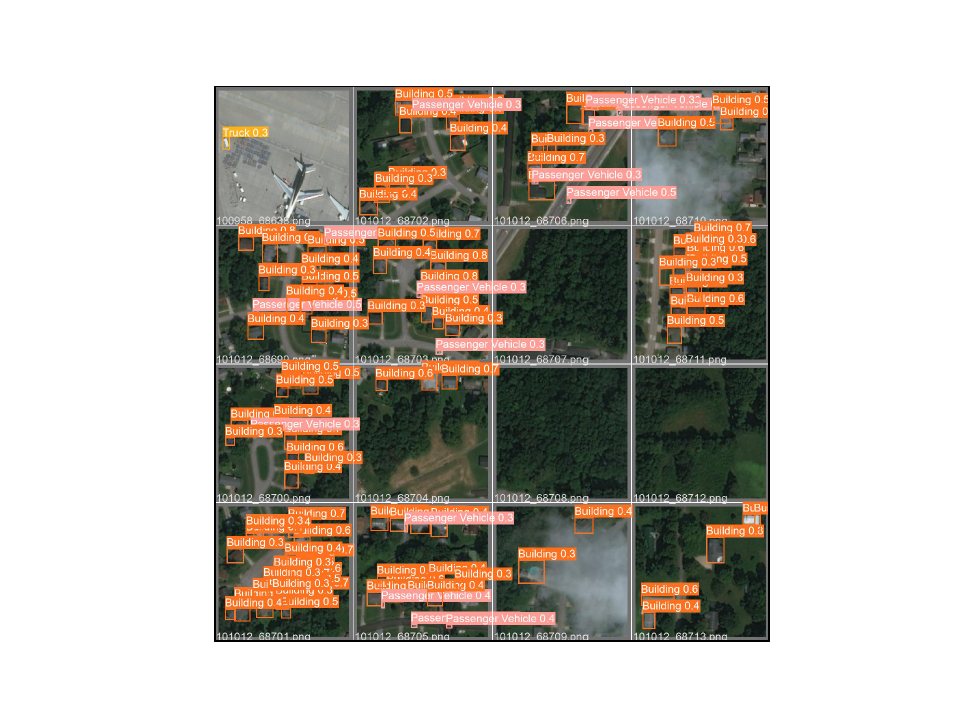}
        \label{Fig:obd-semseg1}
    \end{minipage}%
}
\subfigure[{\color{black}Semantic segmentation}]{
    \begin{minipage}[t]{0.45\linewidth}
        \centering
        \includegraphics[height=3in,trim={60 20 20 0},clip]{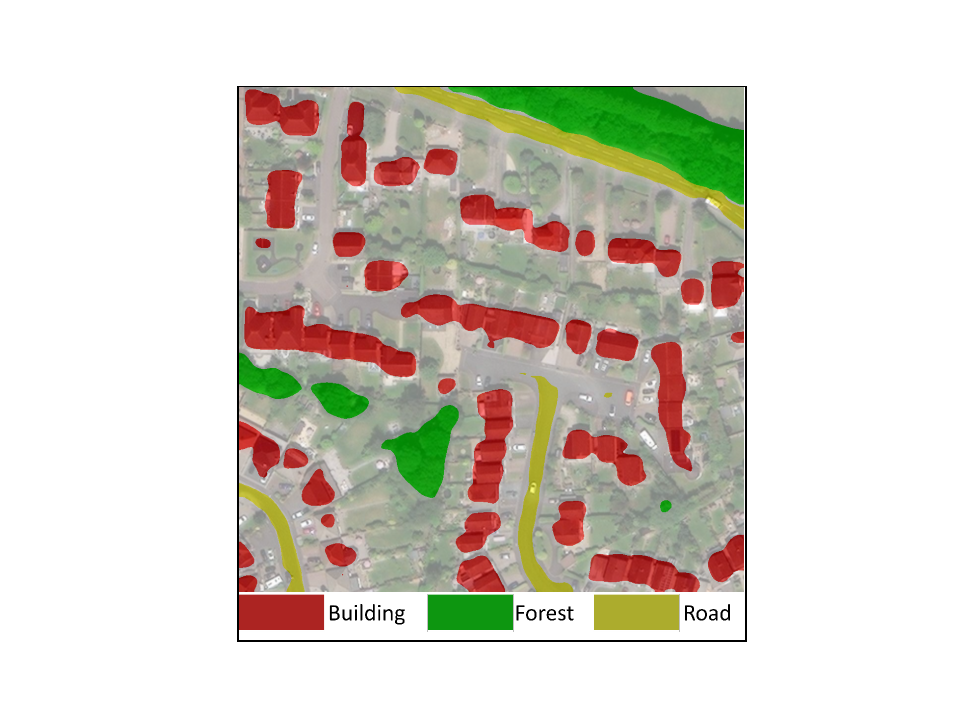}
        \label{Fig:obd-semseg2}
    \end{minipage}%
}
\caption{{\color{black} Visualization of the (a) object detection and (b) semantic segmentation results. (Zoom-in is recommended to visualize the bounding box classes in (a).)}}
\label{Fig:obd-semseg}
\end{figure}

We use state-of-the-art object detection and semantic segmentation models in the computer vision community to extract the visual attributes from satellite imagery. The training datasets, i.e., xView, DOTA v2, and LoveDA datasets, for the deep learning models are frequently used in satellite imagery interpretation tasks. We prepare the training datasets according to the models' requirements and transfer the trained models to the satellite images we collect. {\color{black}Following the evaluation methods in the computer vision community\cite{tan2020efficientdet,wang2021loveda,padillaCITE2020}, we also present the evaluation metrics for the object detection and semantic segmentation models on the evaluation datasets in Table \ref{tab_IOU}. Specifically, we show the accuracy, precision, and recall for all object categories, as well as the mean Average Precision under Intersection over Union (IoU) threshold 0.5 (mAP@0.5) for the object detection models on xView and DOTA v2 datasets, and the accuracy and mean IoU (mIoU) for the semantic segmentation model on LoveDA dataset. Such results guarantee the usefulness and credibility of our produced data.} In addition, to test the robustness of our trained models qualitatively, we randomly select satellite images with their corresponding object detection and semantic segmentation results, which are shown in Figure \ref{Fig:obd-semseg}. We visualize the object detection results in satellite imagery in Figure \ref{Fig:obd-semseg1}, where buildings and passenger vehicles are identified. For the satellite imagery semantic segmentation model, the ViT-Adapter-based model shows high performance in recent semantic segmentation tasks, and the example of segmentation results is shown in Figure \ref{Fig:obd-semseg2}. The results prove the effectiveness of transferring the pre-trained models to our satellite imagery. 

\subsection*{SDG indicators prediction from satellite imagery visual attributes}
\begin{figure}[tb]
\centering
\subfigure[Median household income (log)]{
    \begin{minipage}[t]{0.45\linewidth}
        \centering
        \includegraphics[width=\linewidth]{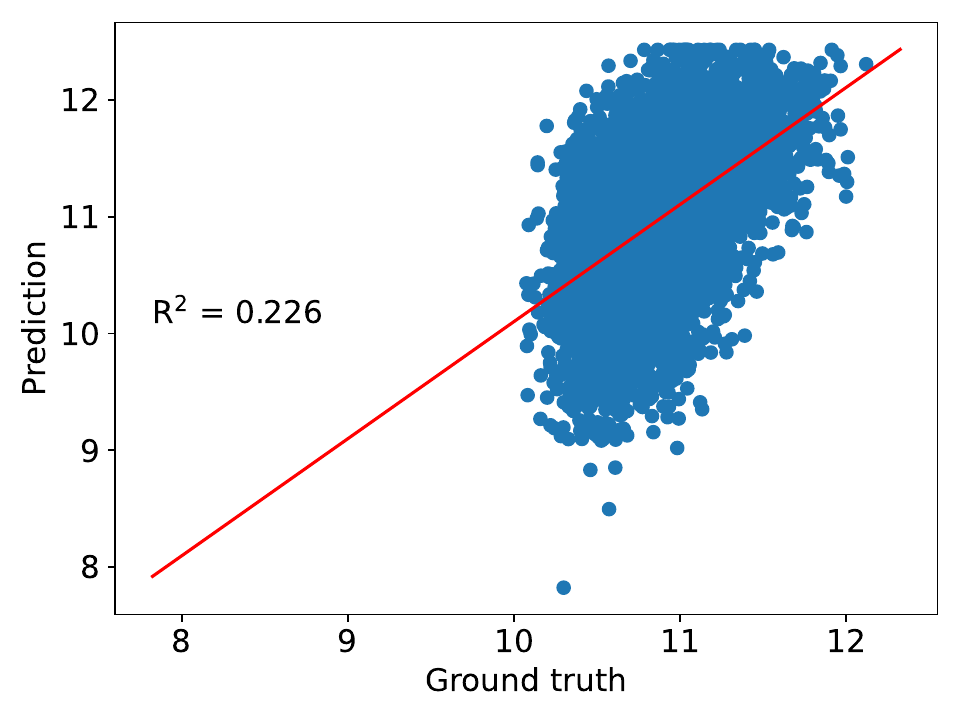}
        \label{FigTempEx1}
    \end{minipage}%
}
\subfigure[Population (log) with no health insurance at all ages]{
    \begin{minipage}[t]{0.45\linewidth}
        \centering
        \includegraphics[width=\linewidth]{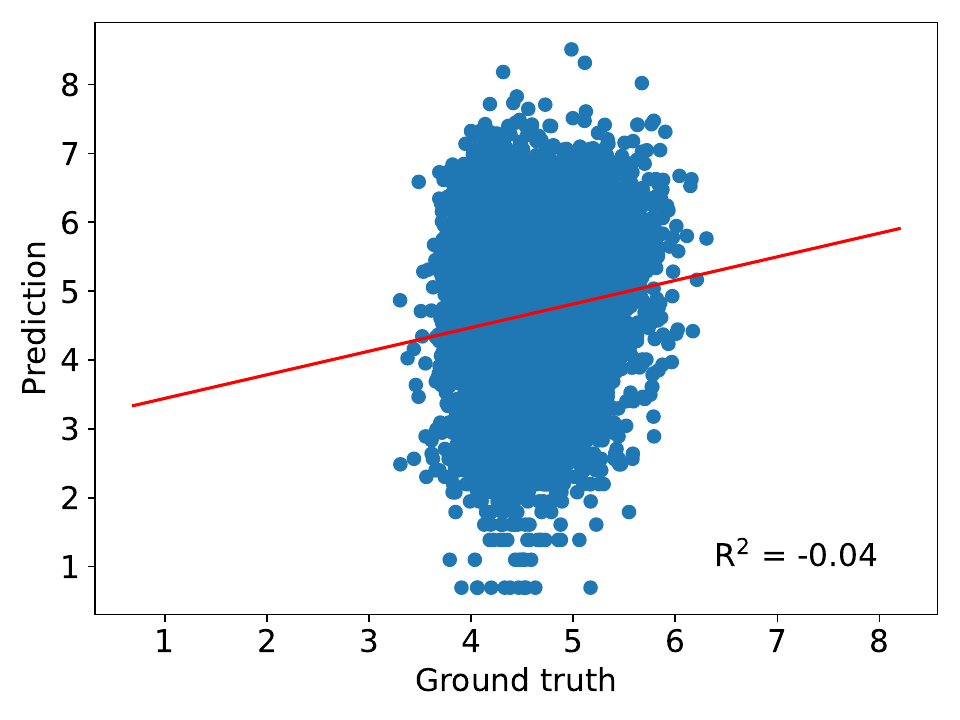}
        \label{FigTempEx2}
    \end{minipage}%
}
\\
\subfigure[Population (log) that graduated from high school]{
    \begin{minipage}[t]{0.45\linewidth}
        \centering
        \includegraphics[width=\linewidth]{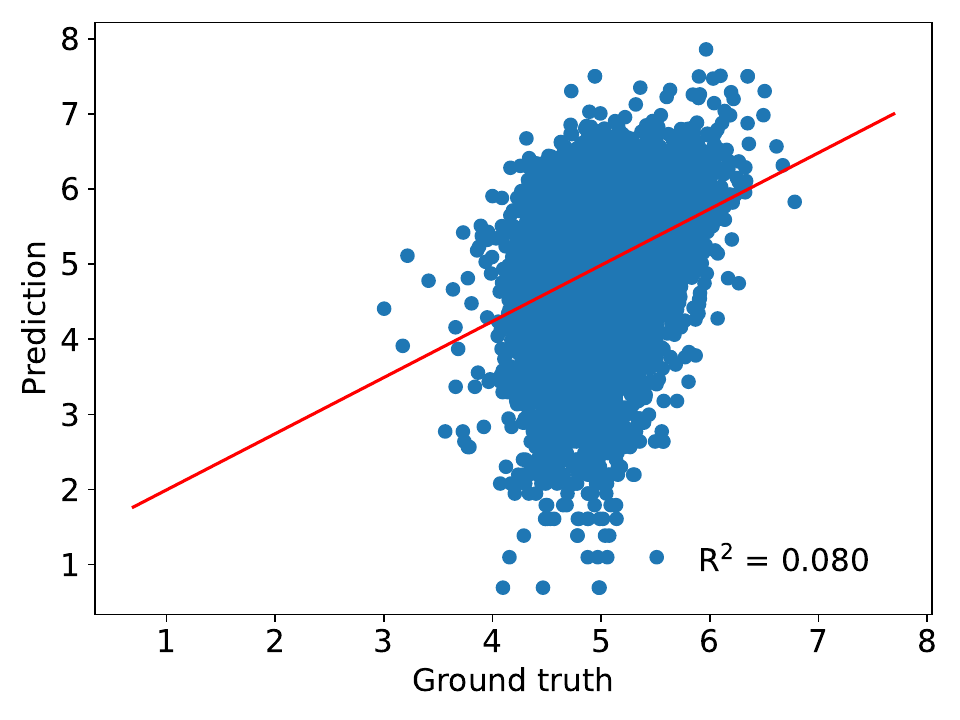}
        \label{FigTempEx3}
    \end{minipage}%
}
\subfigure[POI density (log)]{
    \begin{minipage}[t]{0.45\linewidth}
        \centering
        \includegraphics[width=\linewidth]{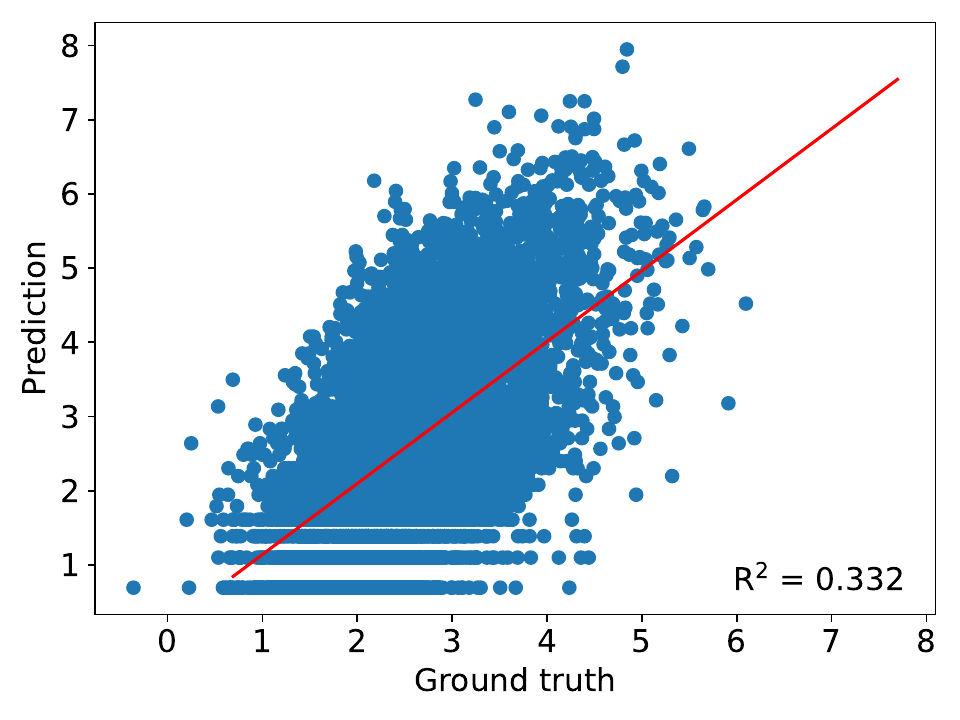}
        \label{FigTempEx4}
    \end{minipage}%
}

\caption{{\color{black} SDG indicators prediction results from satellite imagery visual attributes with GBDT at the CBG level in 2018. (a) Median household income (log), (b) Population (log) with no health insurance at all ages, (c) Population (log) that graduated from high school, and (d) POI density (log).}}
\label{FigPre_GBDT}
\end{figure}

{\color{black} Visual information in satellite imagery correlate with income/daily consumption \cite{ayush2020generating, jean2016combining,han2020lightweight}, commercial {\color{black}activity} \cite{wang2018urban}, education level\cite{head2017can}, and health outcome \cite{head2017can,levy2021using}. Therefore, we validate the possibility of inferring SDG indicators from corresponding satellite imagery. Specifically, for each CBG, the visual attributes (see Table \ref{tab:obdss}) of satellite imagery are fed into a regression model to infer the SDG indicators. We select median household income, population with no health insurance at all ages, population that graduated from high school, and POI density in 2018 as the indicators for poverty, health status, education, and commercial activity, respectively. We experiment on whether those indicators can be inferred from the satellite images by applying Gradient Boosting Decision Trees (GBDT) \cite{friedman2001greedy} on the satellite imagery visual attributes and the indicators selected above as output at the CBG level. The ground truth data for training and validation of the regression models are the collected SDG indicators in our dataset. We randomly split the 100 cities into 80 training cities, and 20 validation cities, and thus all the CBGs in one city are grouped into the same fold.} {\color{black} The regression results are shown in Figure \ref{FigPre_GBDT}, where we can see that the coefficient of determination\cite{lewis2015applied} $R^2$ of the predicted median household income and POI density with regard to the ground truth are higher than the $R^2$ for health and education indicators. Specifically, GBDT has a prediction performance of $R^2$ reaching about $0.22$ and $0.33$ for household median income and POI density, respectively. These results are consistent with the findings in previous research\cite{ayush2020generating, jean2016combining,han2020lightweight,wang2018urban} that socioeconomic status can be inferred from satellite imagery, confirming the validity of the provided dataset and demonstrating the potential to monitor SDGs from satellite imagery. While the education and health indicators are predicted with low precision, which encourages dataset users for future enhancement. At present, most research on predicting socioeconomic status from satellite imagery focuses on income/poverty (SDG 1) and commercial activity (POI density in SDG 11). The studies for inferring regional health or education status are very few, and the performance of relevant prediction models is much lower than the performance of income prediction (see Figure \ref{FigPre_GBDT}), which makes the health and education-related SDG monitoring a promising research direction in the future.}


\section*{Usage Notes}
This study aims to provide a long-term and multi-scale dataset in cities covering the satellite imagery attributes and SDG indicators for urban policymakers and researchers to advance SDG monitoring. Specifically, the satellite imagery attributes in our dataset can be used as input for proposing machine learning models to predict the SDG indicators. Moreover, the SDG data in our dataset can also provide insights into how SDGs evolve in time or scale. Since our dataset contains various aspects of cities, we recommend the following potential research applications: introducing new methods for predicting poverty/income, health, education, inequality, and living environment status of people in cities from long-term or multi-scale satellite images. Researchers are also encouraged to discover the underlying relationship between various SDG progresses and satellite images in cities. 

The dataset files at the CBG level has about $400,000$ lines of data, which might take a long time to load in Excel. Thus, we recommend loading the data with a Python script that can handle large datasets. 




\section*{Code availability}

The Python codes to collect, process, and plot the dataset as well as the supplementary files for this study are publicly available through the GitHub repository {\href{https://github.com/axin1301/Satellite-imagery-dataset} {(https://github.com/axin1301/Satellite-imagery-dataset)}}. Detailed instruction for the running environment, file structure, and codes is available in the repository.




\section*{Acknowledgements}

In this work, we collect ACS data from the United States Census Bureau and SafeGraph Open Census Data, nighttime light data from EOG, and population data from WorldPop project. We also use OpenStreetMap data under the Open Data Commons Open Database License 1.0. The satellite imagery is collected from Esri under the Esri Master License Agreement. Moreover, we use the publicly available object detection and semantic segmentation datasets (xView, DOTA v2, and LoveDA) and codes for deep learning models (YOLO and Vision Transformer Adapter) in this work. We acknowledge these publicly available data sources and codes for supporting this study.

\section*{Author contributions statement}
Yong Li, Tong Li, Yu Liu, Yanxin Xi, and Pan Hui contributed to conceptualizing the study. Yanxin Xi acquired raw data, produced the dataset, and plotted the figures. Yunke Zhang collected the data partly. Yu Liu and Yanxin Xi contributed to the drafting of the manuscript. All authors revised the manuscript.

\section*{Competing interests}
The authors declare no competing interests.




\section*{Figures \& Tables}

\clearpage

\begin{table}[tb]
\renewcommand\arraystretch{1.5}
\centering
\caption{\label{cityname} City-of-interests and corresponding population in our dataset.}
\begin{tabular}{|c|c||c|c||c|c||c|c|}
\hline
\textbf{City} & \textbf{Population} &\textbf{City} & \textbf{Population} &\textbf{City} & \textbf{Population }&\textbf{City} & \textbf{Population}\\
\hline
New York & 8,467,513 &\makecell[c]{Los\\ Angeles} & 3,849,306 &Chicago & 2,696,561 &Houston &2,287,047\\
\hline
Phoenix &1,624,539 &Philadelphia&1,576,251 &\makecell[c]{San \\Antonio} &1,451,863 &San Diego &1,381,600 \\
\hline
Dallas &1,288,441 &San Jose & 983,530 &Austin&964,000 &Jacksonville& 954,624 \\
\hline
Fort Worth &940,437 &Columbus & 907,310&Indianapolis &882,327 & Charlotte&879,697 \\
\hline
\makecell[c]{San\\ Francisco} &815,201 &Seattle & 733,904&Denver & 711,463&Oklahoma &687,691\\
\hline
\makecell[c]{Nashville \\-Davidson}& 678,845&El Paso &678,422 &Washington & 670,050&Boston & 654,281\\
\hline
Las Vegas &646,776&Portland&642,218 &Detroit &632,589 &\makecell[c]{Louisville \\-Jefferson} &628,577 \\
\hline
Memphis & 628,118&Baltimore &576,498&Milwaukee& 569,326&Albuquerque & 562,591\\
\hline
Fresno &544,500 &Tucson & 543,215&Sacramento &525,028&Mesa& 509,492\\
\hline
Kansas &508,415 &Atlanta &496,480 &Omaha &487,299 &\makecell[c]{Colorado\\ Springs} &483,969\\
\hline
Raleigh&469,502 &\makecell[c]{VirGinia \\Beach} &457,672 &Long Beach & 456,063&Miami &439,906 \\
\hline
Oakland & 433,797&Minneapolis& 425,338& Tulsa&411,905 &Bakersfield &407,581 \\
\hline
Wichita & 395,707&Arlington & 392,802&Aurora &389,675& Tampa&387,037 \\
\hline
New Orleans &376,971 &Cleveland &368,006 &Anaheim & 345,935&Henderson &322,202 \\
\hline
Stockton&322,107 &\makecell[c]{Lexington \\-Fayette} &321,793 &\makecell[c]{Corpus\\ Christi} & 317,768&Riverside & 317,257\\
\hline
Santa Ana &309,468 & Orlando&309,193 &Irvine &309,014 &Cincinnati &308,913  \\
\hline
Newark &307,216&St. Paul&307,176  & Pittsburgh&300,454 &Greensboro &298,250 \\
\hline
St. Louis &293,310 &Lincoln & 292,648&Plano&287,037 & Durham& 285,439 \\
\hline
Jersey & 283,943&Chandler &279,445 &\makecell[c]{Chula\\ Vista}&277,211&Buffalo& 276,804 \\
\hline
 \makecell[c]{North \\Las Vegas}& 274,146&Gilbert & 273,138&Madison &269,162 &Reno &268,843\\
\hline
Toledo&268,504 & \makecell[c]{Fort \\Wayne}&263,814 &Lubbock &260,990 &St. Petersburg & 258,214 \\
\hline
Laredo & 258,014&Irving&254,190 &Chesapeake&251,269 &\makecell[c]{Winston\\ -Salem} &250,337   \\
\hline
Glendale &249,627&\makecell[c]{Enterprise\\ CDP} &245,286& Scottsdale&242,754 &Garland&241,870   \\
\hline
Boise &237,457&Norfolk &235,089 &\makecell[c]{Arlington\\ CDP} &232,965& Spokane&229,065  \\
\hline
Fremont &227,523 &Richmond & 226,604&\makecell[c]{Santa\\ Clarita} &224,588 &\makecell[c]{San\\ Bernardino} &222,194\\
\hline
\end{tabular}
\end{table}

\begin{table}[tb]
\renewcommand\arraystretch{1.5}
\centering
\caption{\label{tab2}Example of the geographic lookup table between cities and CBGs in the produced dataset.}
\begin{tabular}{|c|c|c|c|c|}
\hline
\textbf{City Name} & \textbf{City GeoID} & \textbf{Track Code} &\textbf{Block Group} &\textbf{CBG Code}\\
\hline
New York & 1600000US3651000 & 17500& 2 & 360050175002 \\
\hline
New York & 1600000US3651000 & 14100& 1 & 360050141001 \\
\hline
New York & 1600000US3651000 & 14500& 2 & 360050145001 \\
\hline
\multicolumn{5}{|c|}{$\cdots$}\\
\hline
Los Angeles & 1600000US0644000 & 216401&1 & 060372164011 \\
\hline
Los Angeles & 1600000US0644000 & 216401&2 & 060372164012 \\
\hline
Los Angeles & 1600000US0644000 & 216402&1 & 060372164021 \\
\hline
\multicolumn{5}{|c|}{$\cdots$}\\
\hline
\end{tabular}
\end{table}

\begin{table}[tb]
\renewcommand\arraystretch{1.5}
\centering
\caption{\label{tab:obdss} Visual attributes extracted from satellite images.}
\begin{tabular}{|c|c|}
\hline
\textbf{Objects Detected} & \makecell[c]{ passenger vehicles, swimming pools, planes, airports, trucks, \\railway vehicles, ships, engineering vehicles, bridges, roundabouts,\\  vehicle lots, soccer fields, basketball courts, ground track fields, \\baseball diamonds, tennis courts, buildings (number of buildings)}\\
\hline
\textbf{Land Cover Semantics} & road, forest, building (pixel percentage), water, barren, agriculture, background \\
\hline
\end{tabular}
\end{table}

\begin{table}[tb]
\renewcommand\arraystretch{1.5}
\centering
\caption{\label{tab_corr} {\color{black}Relationship between the SDG targets and the selected indicators.}}
\begin{threeparttable}
\begin{tabular}{|c|c||c|c|}
\hline
{\color{black}SDG Target} &  {\color{black}Indicator} &  {\color{black}SDG Target}  &  {\color{black}Indicator}\\
\hline
{\color{black}1.2}  & \textbf{ {\color{black}Population Below Poverty}}  &{\color{black} 1.2 }&\textbf{{\color{black}Population Above Poverty}} \\
\hline
{\color{black}1.2}  & \textbf{{\color{black}\makecell[c]{ Population With A Ratio \\Of Income To Poverty Level\\ Under 0.5}}}  &{\color{black} 1.2}&\textbf{{\color{black}\makecell[c]{ Population With A Ratio \\Of Income To Poverty Level\\ 0.5 to 0.99}}} \\
\hline
 {\color{black}1.4 } & \textbf{ {\color{black}Median Household Income}} &  &  \\
\hline
 {\color{black}3.8 } & \textbf{{\color{black}\makecell[c]{Population With No Health \\Insurance Under 18}}}  & {\color{black}3.8} & \textbf{{\color{black}\makecell[c]{Population With No Health\\ Insurance Between 18 To 34}}} \\
\hline
 {\color{black}3.8}  &\textbf{{\color{black}\makecell[c]{Population With No Health\\ Insurance Between 35 To 64}}}  &  {\color{black}3.8}  &\textbf{{\color{black}\makecell[c]{Population With No Health\\ Insurance Over 65 Years Old} }} \\
\hline
{\color{black}3.8}  &\textbf{{\color{black}\makecell[c]{Civilian Noninstitutionalized Population}}}   &    & \\
\hline
 {\color{black}4.1}  & \textbf{{\color{black}\makecell[c]{Population Enrolled In   College} }}  & {\color{black} 4.3 } & \textbf{{\color{black}\makecell[c]{Population That Graduated\\ From High School}}}  \\
\hline
{\color{black} 4.3 } &  \textbf{{\color{black}\makecell[c]{Population With A \\Bachelor's Degree}}}  & {\color{black} 4.3}  &\textbf{{\color{black}\makecell[c]{Population With A \\Master's Degree}}}  \\
\hline
 {\color{black}4.3 } & \textbf{{\color{black}\makecell[c]{Population  With A Doctorate}} } & & \\
\hline
 {\color{black}10.2}  &\textbf{{\color{black}Light Gini} } &{\color{black}10.2 } &\textbf{{\color{black}Income Gini}}   \\
\hline

{\color{black}11.2}  & \textbf{{\color{black}Driving Road Density}}  & {\color{black}11.2}  & \textbf{{\color{black}Cycling Road Density}}  \\
\hline
{\color{black} 11.2}  & \textbf{{\color{black}Walking Road Density} } &{\color{black}11.3 } & \textbf{{\color{black}POI Density}}    \\
\hline
{\color{black}11.3}  &\textbf{{\color{black} Building Density}}    & {\color{black} 11.3}  &\textbf{{\color{black}Land Use}}  \\
\hline
{\color{black}11.3}  &\textbf{{\color{black}Entropy Index} } &  {\color{black}11.3 } &\textbf{{\color{black}Index of Dissimilarity }} \\
\hline
\end{tabular}
    \begin{tablenotes}
    \footnotesize
    {\color{black}
    \item[1] Target 1.2: By 2030, reduce at least by half the proportion of men, women, and children of all ages living in poverty in all its dimensions according to national definitions.
    \item[2] Target 1.4: By 2030, ensure that all men and women, in particular the poor and the vulnerable, have equal rights to economic resources, as well as access to basic services, ownership and control over land and other forms of property, inheritance, natural resources, appropriate new technology and financial services, including microfinance.
    \item[3] Target 3.8: Achieve universal health coverage, including financial risk protection, access to quality essential healthcare services, and access to safe, effective, quality, and affordable essential medicines and vaccines for all.
    \item[4] Target 4.1: By 2030, ensure that all girls and boys complete free, equitable, and quality primary and secondary education leading to relevant and effective learning outcomes.
    \item[5] Target 4.3: By 2030, ensure equal access for all women and men to affordable and quality technical, vocational and tertiary education, including university.
    \item[6] Target 10.2: By 2030, empower and promote the social, economic and political inclusion of all, irrespective of age, sex, disability, race, ethnicity, origin, religion or economic or other status.
    \item[7] Target 11.2: By 2030, provide access to safe, affordable, accessible and sustainable transport systems for all, improving road safety, notably by expanding public transport, with special attention to the needs of those in vulnerable situations, women, children, persons with disabilities and older persons.
    \item[8] Target 11.3: By 2030, enhance inclusive and sustainable urbanization and capacity for participatory, integrated and sustainable human settlement planning and management in all countries.}
    \end{tablenotes}
\end{threeparttable}
\end{table}

\begin{table}[tb]
\renewcommand\arraystretch{1.5}
\centering
\caption{\label{tab_IOU} {\color{black}Evaluation metrics for the object detection and semantic segmentation models.}}
\begin{tabular}{|c|c|c|c|c|c|c|}
\hline
{\color{black}Model} & {\color{black}Dataset} & {\color{black}Accuracy(\%)}  & {\color{black}Precision(\%)} & {\color{black}Recall(\%)}& {\color{black}mAP@0.5(\%)} & {\color{black}mIoU(\%)} \\
\hline
\multirow{2}*{{\color{black}Object Detection}} &{\color{black}xView} & {\color{black}66.7}  & {\color{black}53.2} & {\color{black}37.7}& {\color{black}37.1}  & {\color{black}-} \\
\cline{2-7}
~ & {\color{black}DOTA v2}& {\color{black}47.7}  & {\color{black}77.0} & {\color{black}51.8} & {\color{black}58.2}  & {\color{black}-} \\
\hline
{\color{black}Semantic Segmentation} & {\color{black}LoveDA} & {\color{black}71.1}  & {\color{black}-}& {\color{black}-} & {\color{black}-} & {\color{black}52.7}\\
\hline
\end{tabular}
\end{table}

\begin{table}[tb]
\renewcommand\arraystretch{1.5}
\centering
\caption{\label{tab3}Example of basic geographic statistics of CBGs in the produced dataset.}
\begin{tabular}{|c|c|c|c|c|c|c|}
\hline
\multicolumn{3}{|c}{Geographic Area} &\multicolumn{4}{|c|}{Basic Geographical Statistics}\\
\hline
\textbf{City Name} & \textbf{CBG Code}& \textbf{Year} & \textbf{Population} &\textbf{Area}& \textbf{Centroid} &\textbf{Boundary}\\
\hline
Albuquerque & 350010001071 & 2014&1,965&2.50&(-106.48718, 35.12321)&\makecell[c]{(-106.49789 \\35.13066,\\ -106.49730\\ 35.13067, ...)} \\
\hline
Albuquerque & 350010001071 & 2015&1,992&2.50&(-106.48718, 35.12321)&\makecell[c]{(-106.49789 \\35.13066,\\ -106.49730\\ 35.13067, ...)} \\
\hline
Albuquerque & 350010001071 & 2016&2,020&2.50&(-106.48718, 35.12321)&\makecell[c]{(-106.49789 \\35.13066,\\ -106.49730\\ 35.13067, ...)} \\
\hline
\multicolumn{7}{|c|}{$\cdots$}\\
\hline
\end{tabular}
\end{table}

\begin{table}[tb]
\renewcommand\arraystretch{1.5}
\centering
\caption{\label{tab4}Example of basic geographical statistics of cities in the produced dataset.}
\begin{tabular}{|c|c|c|c|c|c|c|}
\hline
\multicolumn{3}{|c}{Geographic Area} &\multicolumn{4}{|c|}{Basic Geographical Statistics}\\
\hline
\textbf{City Name} & \textbf{City GeoID}& \textbf{Year} & \textbf{Population} &\textbf{Area}& \textbf{Centroid} &\textbf{Boundary}\\
\hline
Albuquerque& 1600000US3502000 & 2014&577,889&489&(-106.64648, 35.10534)&\makecell[c]{(-106.64882 \\35.14807,\\ -106.64878\\ 35.14818, ...)} \\
\hline
Albuquerque& 1600000US3502000 & 2015&585,825&489&(-106.64648, 35.10534)&\makecell[c]{(-106.64882 \\35.14807,\\ -106.64878\\ 35.14818, ...)}\\
\hline
Albuquerque & 1600000US3502000 & 2016&593,571&489&(-106.64648, 35.10534)&\makecell[c]{(-106.64882 \\35.14807,\\ -106.64878\\ 35.14818, ...)}\\
\hline
\multicolumn{7}{|c|}{$\cdots$}\\
\hline
\end{tabular}
\end{table}

\begin{table}[tb]
\renewcommand\arraystretch{1.5}
\centering
\caption{\label{sat}Example of visual attributes from satellite images at the CBG level in the produced dataset.}
 \begin{threeparttable}
\begin{tabular}{|c|c|c|c|c|c|c|c|c|}
\hline
\multicolumn{2}{|c}{Geographic Area}&\  &\multicolumn{3}{|c}{Objects Detected}&\multicolumn{3}{|c|}{\makecell[c]{Land Cover\\ Semantics}}\\
\hline
\textbf{City Name} & \textbf{CBG Code}& \textbf{Year} & \textbf{\makecell[c]{Passenger\\ Vehicle} }&\textbf{\makecell[c]{Swimming \\ Pool}}& \textbf{\makecell[c]{$\cdots$}} &\textbf{\makecell[c]{Road}}&\textbf{\makecell[c]{Forest} }&\textbf{\makecell[c]{$\cdots$}}\\
\hline
\makecell[c]{Albuquerque} & 350010001071 & 2014 &3,176&665&$\cdots$&0.0391&0.395&$\cdots$ \\
\hline
\makecell[c]{Albuquerque} & 350010001072 & 2014 &1,465&639&$\cdots$&0.0712&0.00979&$\cdots$ \\
\hline
\makecell[c]{Albuquerque} & 350010001081 & 2014 &2,150&527&$\cdots$&0.124&0.000537&$\cdots$ \\
\hline
\multicolumn{9}{|c|}{$\cdots$}\\
\hline
\end{tabular}
    \begin{tablenotes}
    \footnotesize
    \item[1] Objects detected consist of 17 columns, as demonstrated in Table \ref{tab:obdss}.
    \item[2] Land cover semantics consist of 7 columns, as demonstrated in Table \ref{tab:obdss}.
    \end{tablenotes}
\end{threeparttable}
\end{table}

\begin{table}[tb]
\renewcommand\arraystretch{1.5}
\centering
\caption{\label{tab5}Example of indicators for SDG 1 at the CBG level in the produced dataset.}
\begin{tabular}{|c|c|c|c|c|c|c|c|c|c|}
\hline
\multicolumn{3}{|c}{Geographic Area} &\multicolumn{5}{|c|}{SDG 1 (No Poverty)}\\
\hline
\textbf{City Name} & \textbf{CBG Code}& \textbf{Year} & \textbf{\makecell[c]{Median\\ Household\\ Income} } &\textbf{\makecell[c]{Population \\Above\\ Poverty}}&\textbf{\makecell[c]{Population \\Below\\ Poverty} }&\textbf{\makecell[c]{Population \\With A \\Ratio Of \\Income To\\ Poverty\\ Level\\ Under 0.5}}&\textbf{\makecell[c]{Population \\With A \\Ratio Of\\ Income To\\ Poverty\\ Level\\ 0.5 to 0.99}}\\
\hline
Albuquerque & 350010001071 & 2014 &70,625  &1,358  &169 &61 &108  \\
\hline
Albuquerque& 350010001071 & 2015 &79,276  &1,293  &133 &57 &76 \\
\hline
Albuquerque & 350010001071 & 2016 &95,000  &1,531  &151 &81  &70\\
\hline
\multicolumn{8}{|c|}{$\cdots$}\\
\hline
\end{tabular}
\end{table}

\begin{table}[tb]
\renewcommand\arraystretch{1.5}
\centering
\caption{\label{tab6}Example of indicators for SDG 3 at the CBG level in the produced dataset.}
\begin{tabular}{|c|c|c|c|c|c|c|c|}
\hline
\multicolumn{3}{|c}{Geographic Area} &\multicolumn{5}{|c|}{SDG 3 (Good Health and Well-being)}\\
\hline
\textbf{City Name} & \textbf{CBG Code}& \textbf{Year} & \textbf{\makecell[c]{Civilian\\ Noninstitutionalized \\Population} }&\textbf{\makecell[c]{Population \\With No \\Health \\Insurance \\Under 18}}& \textbf{\makecell[c]{Population \\With No \\Health \\Insurance\\Between \\18 To 34}} &\textbf{\makecell[c]{Population \\With No \\Health \\Insurance\\Between \\35 To 64}}&\textbf{\makecell[c]{Population \\With No \\Health \\Insurance\\ Over 65\\ Years Old} }\\
\hline
Albuquerque & 350010001071 & 2014  &1,682 &0  &22 &8  &0 \\
\hline
Albuquerque & 350010001071 & 2015   &1,678  &0  &19  &25  &0 \\
\hline
Albuquerque & 350010001071 & 2016  & 1,532  &0  &46  &22  &0 \\
\hline
\multicolumn{8}{|c|}{$\cdots$}\\
\hline
\end{tabular}
\end{table}

\begin{table}[tb]
\renewcommand\arraystretch{1.5}
\centering
\caption{\label{tab7}Example of indicators for SDG 4 at the CBG level in the produced dataset.}
\begin{tabular}{|c|c|c|c|c|c|c|c|}
\hline
\multicolumn{3}{|c}{Geographic Area} &\multicolumn{5}{|c|}{SDG 4 (Quality Education)}\\
\hline
\textbf{City Name} & \textbf{CBG Code}& \textbf{Year} & \textbf{\makecell[c]{Population \\Enrolled \\In College} }&\textbf{\makecell[c]{Population \\ That \\ Graduated \\From\\ High School}}&\textbf{\makecell[c]{Population \\With A \\Bachelor's\\ Degree}} &\textbf{\makecell[c]{Population \\With A \\Master's\\ Degree}}&\textbf{\makecell[c]{Population \\ With A \\Doctorate} }\\
\hline
Albuquerque & 350010001071 & 2014 &77 &171 &267 &242 &108 \\
\hline
Albuquerque & 350010001071 & 2015  &50 &140 &233 &259 &97 \\
\hline
Albuquerque & 350010001071 & 2016  &77 &207 &276 &250 &108 \\
\hline
\multicolumn{8}{|c|}{$\cdots$}\\
\hline
\end{tabular}
\end{table}

\begin{table}[tb]
\renewcommand\arraystretch{1.5}
\centering
\caption{\label{tab8}{\color{black}Example of indicators for SDG 10 at the CBG level in the produced dataset.}}
\begin{threeparttable}
\begin{tabular}{|c|c|c|c|c|}
\hline
\multicolumn{3}{|c}{\color{black}{Geographic Area} }&\multicolumn{2}{|c|}{{\color{black}SDG 10 (Reduced Inequalities)}}\\
\hline
\textbf{{\color{black}City Name}} & \textbf{{\color{black}CBG Code}}& \textbf{{\color{black}Year}} & \textbf{\makecell[c]{{{\color{black}Light Gini}}} }&\textbf{\makecell[c]{{{\color{black}Income Gini}}}}\\
\hline
{\color{black}Albuquerque} & {\color{black}350010001071} & {\color{black}2014} &{\color{black}0.761} &{\color{black}-} \\
\hline
{\color{black}Albuquerque} & {\color{black}350010001071} & {\color{black}2015}  &{\color{black}0.404} &{\color{black}-} \\
\hline
{\color{black}Albuquerque} & {\color{black}350010001071} & {\color{black}2016}  &{\color{black}0.392} &{\color{black}-} \\
\hline
\multicolumn{5}{|c|}{$\cdots$}\\
\hline
\end{tabular}
    \begin{tablenotes}
    \footnotesize
    {\color{black}\item[1] The income Gini at the CBG level is not available in ACS data.}
    \end{tablenotes}
\end{threeparttable}
\end{table}

\begin{table}[tb]
\renewcommand\arraystretch{1.5}
\centering
\caption{\label{tab9}{\color{black}Example of indicators for SDG 11 at the CBG level in the produced dataset.}}
\begin{threeparttable}
\begin{tabular}{|c|c|c|c|c|c|c|c|c|}
\hline
\multicolumn{3}{|c}{Geographic Area} &\multicolumn{6}{|c|}{SDG 11 (Sustainable Cities and Communities)}\\
\hline
\textbf{City Name} & \textbf{CBG Code}& \textbf{Year} & \textbf{\makecell[c]{Building\\ Density} }&\textbf{\makecell[c]{Driving/\\Cycling/\\Walking\\ Road\\ Density}}&\textbf{\makecell[c]{ POI\\ Density} }&\textbf{\makecell[c]{ Land Use}} &\textbf{{\color{black}\makecell[c]{Index of \\Dissimilarity}}} & \textbf{{\color{black}\makecell[c]{Entropy\\ Index}}}\\
\hline
Albuquerque & 350010001071 & 2014 &20.3 &8.28/9.63/9.87 &1.59&0/0/0/33.7\%&{\color{black}-} &{\color{black}0.702}\\
\hline
Albuquerque & 350010001071 & 2015 &39.9 &8.87/11.6/11.9 &1.99&0/0/0/33.7\%&{\color{black}-} &{\color{black}0.674} \\
\hline
Albuquerque & 350010001071 & 2016 &39.9 &8.71/11.5/11.7&1.99&0/0/0/33.7\%&{\color{black}-} &{\color{black}0.738} \\
\hline
\multicolumn{9}{|c|}{$\cdots$}\\
\hline
\end{tabular}
    \begin{tablenotes}
    \footnotesize
    \item[1] Land Use consists of 4 columns: Commercial/ Industrial/ Construction/ Residential.
    {\color{black}\item[2] Index of Dissimilarity has 6 columns: White-Black/ White-Asian/ White-Hispanic/ Black-Asian/ Black-Hispanic/ Asian-Hispanic.}
    {\color{black}\item[3] In CBG, there is no data for proportions of the population with different races or ethnicity, so the index of dissimilarity for CBG is not provided.}
    \end{tablenotes}
\end{threeparttable}
\end{table}


\end{document}